# Cycle Consistency-based Uncertainty Quantification of Neural Networks in Inverse Imaging Problems

Luzhe Huang[1,2,3], Jianing Li[4], Xiaofu Ding[5], Yijie Zhang[1,2,3], Hanlong Chen[1,2,3], Aydogan Ozcan[1,2,3,6,*]

[1] Electrical and Computer Engineering Department, University of California, Los Angeles, CA 90095, USA

[2] Bioengineering Department, University of California, Los Angeles, CA 90095, USA

[3] California NanoSystems Institute (CNSI), University of California, Los Angeles, CA 90095, USA.

[4] Computer Science Department, University of California, Los Angeles, CA, 90095, CA

[5] Department of Mathematics, University of California, Los Angeles, CA, 90095, CA

[6] David Geffen School of Medicine, University of California Los Angeles, Los Angeles, CA 90095, USA

Corresponding author: ozcan@ucla.edu

**Abstract**

Uncertainty estimation is critical for numerous applications of deep neural networks and draws growing attention from researchers. Here, we demonstrate an uncertainty quantification approach for deep neural networks used in inverse problems based on cycle consistency. We build forward-backward cycles using the physical forward model available and a trained deep neural network solving the inverse problem at hand, and accordingly derive uncertainty estimators through regression analysis on the consistency of these forward-backward cycles. We theoretically analyze cycle consistency metrics and derive their relationship with respect to uncertainty, bias, and robustness of the neural network inference. To demonstrate the effectiveness of these cycle consistency-based uncertainty estimators, we classified corrupted and out-of-distribution input image data using some of the widely used image deblurring and super-resolution neural networks as testbeds. The blind testing of our method outperformed other models in identifying unseen input data corruption and distribution shifts. This work provides a simple-to-implement and rapid uncertainty quantification method that can be universally applied to various neural networks used for solving inverse problems.

## 1. Introduction

In the past decades, deep learning achieved enormous advances, demonstrating unprecedented performance, and significantly impacted numerous research fields, e.g., data mining[1], natural language processing[2], and computer vision (CV)[3]. For example, the widespread applications of deep neural networks (DNNs) in CV fields can be seen in autonomous driving [4], biomedical imaging and microscopy[5,6]. However, there is still a strong need for further improvements regarding DNN's reliability since incorrect network inference can lead to wrong decisions and jeopardize their uses in critical real-world applications. Most existing deep learning models cannot provide reliable uncertainty quantification (UQ) for their predictions to distinguish input data distribution shifts during the test stage[7,8] or to detect adversarial attacks[9,10].

The sources of uncertainty are categorized into data (aleatoric) uncertainty and model (epistemic) uncertainty[11,12]. Data uncertainty comes from the inherent errors and random perturbations of the measurement system and data acquisition process, e.g., measurement noise[13] or image distortions[14]. On the other hand, model uncertainty usually results from model imperfections caused by DNN architecture design and stochastic training procedure, as well as generalization errors on unknown data distributions[15]. Some of the existing deep learning methods solving inverse imaging problems attempt to avoid data uncertainty caused by perturbations in the physical systems, e.g., movements/misalignments of the imaging elements or the objective lens, through training techniques involving specialized loss functions, training protocols or image data augmentation[16–18]. However, despite these existing efforts to alleviate data uncertainty in the network training, the joint effects of uncontrollable perturbations and model uncertainty make it nearly impossible to eliminate inference uncertainty for various practical applications. Thus, estimating the uncertainty of network inference has drawn increasing interest among researchers. The most common uncertainty quantification approaches are based on Bayesian techniques[15,19–24]. Bayesian uncertainty quantification has been applied in inverse imaging problems[19,20,25], especially in the field of biomedical imaging[26–30]. For example, Bayesian methods were integrated into neural network training procedures to provide uncertainty estimation[21,31,32]. Other techniques were also demonstrated for UQ in inverse problems, such as ensemble learning[33–35] and function priors[36]. However, most of these existing UQ methods require large amounts of training data, additional network modules, or significant modifications to the training process.

In this work, we introduce a simple yet powerful UQ method to quantitatively measure the uncertainty of neural network outputs in solving inverse problems and automatically detect input data corruption and distribution shifts that have never been seen before. As shown in Fig. 1, this approach is based on the execution of forward-backward cycles using the physical forward model and an accordingly trained neural network in an iterative manner. Through a regression-based method, we use these iterative back-and-forth cycles to estimate the neural network's inference uncertainty and derive uncertainty estimators utilizing cycle consistency as a quantitative measure (Fig. 1(b)). Furthermore, we theoretically establish the relationship of these cycle consistency metrics with respect to the uncertainty, bias, and robustness of the neural network inference. Based on these fitted uncertainty estimators, a linear regression model was established to accurately predict the uncertainty for each image inference, and a simple and fast binary classifier was trained to successfully distinguish out-of-distribution (OOD) data where the neural network would normally produce significant generalization errors (Fig. 1(c)) – i.e., helping us avoid catastrophic errors in the network inference.

This UQ method broadly applies to various deep learning-based solutions to inverse problems. To demonstrate its effectiveness, we validated this UQ approach on two tasks used as our testbeds: (1) corrupted input detection for an image deblurring neural network and (2) OOD data detection for an image super-resolution (SR) neural network. Compared to existing UQ methods, our method is simpler as it does not require any modification to the neural network architecture, training or testing processes, and can be directly applied to various trained networks as a universal approach. In addition, our method is independent of the distributions of the OOD data during the test stage. Through our testbeds, we demonstrated that the OOD data classifier could be trained on a limited set of outliers generated by introducing, on purpose, noise into in-distribution (ID) image data to successfully generalize to OOD test image data perturbed by unseen factors, e.g., different noise levels, noise distributions, and different object classes never seen before. Our results reveal that the presented approach, with its cycle-consistency metrics, outperforms existing OOD detection methods across multiple inference tasks; specifically, our approach outperformed other supervised deep neural network-based methods and has significantly less model complexity and can be trained much faster, in ≤ 1 sec. This cycle-consistency-based UQ framework provides a simple yet powerful approach to rapidly quantify output uncertainty in various fields, where neural network inference is used to solve inverse problems.

## 2. Results

### 2.1 Theoretical bounds of cycle consistency

In most imaging and sensing tasks, low-dimensional measurements $x \in R^N$ about a ground truth high-dimensional signal $y \in R^M$ are captured through a non-linear measurement system. The forward process of such a measurement can be formulated by[5]

$$x = f \circ h(y) + n \qquad (1)$$

, where $f$ and $h$ represent the sampling process and the transfer function of the measurement system, respectively. ∘ represents function composition, and $n$ is the random noise present in the measurement system. In most cases, we can assume a linear transfer function and simplify its implementation using the multiplication of the input with a transfer matrix $H \in R^{N \times M}$, representing the physical forward model.

Without loss of generality, we select inverse problems in imaging as our testbed in this manuscript. A network is trained on a dataset $\mathcal{D} = \{(x, y), x \in X, y \in Y\}$, where $X$ and $Y$ denote the domains of the input (measurements) and target images (ground truth), respectively. We denote a trained network to estimate the ground truth image from the measurements as $g_\theta$ parameterized by $\theta$ such that:

$$y_0 = g_\theta(x) = y + \varepsilon_0 \qquad (2)$$

$\varepsilon_0$ represents the error (uncertainty) between the network output $y_0$ and the ground truth $y$.

In general, such uncertainty is hard to estimate without the knowledge of the distribution of the ground truth signals during the test stage. To address this issue, we build forward-backward cycles between the input and target domains, where the uncertainty accumulates iteratively and can be effectively estimated (see Fig. 1). These cycles are built by sequentially passing the images through the neural network ($g_\theta$) and the deterministic physical forward model, i.e.,

$$\begin{cases} y_0 = g_\theta(x) \\ x_n = f \circ h(y_{n-1}) \\ y_n = g_\theta(x_n) \end{cases} \quad (3)$$

where $n = 1, 2, \cdots$ is the cycle index. Through these cycles, we get two image sequences, $\{x, x_1, x_2, \cdots, x_n\}$ and $\{y_0, y_1, \cdots, y_n\}$, in the input and target domains, respectively.

Next, we theoretically show that the cycle consistency, defined as the difference between two adjacent outputs $\|y_{n+1} - y_n\|$, has an upper bound proportional to $\|\varepsilon_0\|$ with two assumptions about the forward physical process and the trained model:

1. The trained model is unbiased on the data distribution $\mathcal{D} = \{(x, y), x \in X, y \in Y\}$, i.e.,

$$y = g_\theta(f \circ h(y)), \forall y \in Y \quad (4)$$

2. The deterministic forward-backward cycle satisfies Lipschitz continuity with a constant $L$ around a neighbor of $y$ containing $y_0, y_1, \cdots y_N$.

$$\|g_\theta(f \circ h(z_1)) - g_\theta(f \circ h(z_2))\| \leq L\|z_1 - z_2\|, \forall z_1, z_2 \in N_y \quad (5)$$

$N_y$ is the neighbor of $y$, and $z_i \in N_y, i = 0, 1, 2, \cdots, N$. $N$ is the number of maximum cycles. The Lipschitz continuity of common neural network components, e.g., linear and convolutional layers, was proven, and efficient estimation of the corresponding Lipschitz constants has been thoroughly studied in the literature[37–39].

To quantify the uncertainty $\varepsilon_0$ *without* any access to the ground truth $y$, $y_0$ is passed through the forward-backward cycles, with the uncertainty accumulating gradually. We can derive the following recursive relationship for the differences between adjacent outputs:

$$\|y_{n+1} - y_n\| = \|g_\theta(f \circ h(y_n)) - g_\theta(f \circ h(y_{n-1}))\| \leq L\|y_n - y_{n-1}\|, \forall n \geq 1 \quad (6)$$

By induction, the difference can be further bounded by $\|\varepsilon_0\|$:

$$\begin{aligned} \|\Delta y_n\| = \|y_n - y_{n-1}\| &\leq L\|y_{n-1} - y_{n-2}\| \leq L^{n-1}\|y_1 - y_0\| \\ &= L^{n-1}\|g_\theta(f \circ h(y + \varepsilon_0)) - y - \varepsilon_0\| \\ &\leq L^{n-1}(\|g_\theta(f \circ h(y + \varepsilon_0)) - y\| + \|\varepsilon_0\|) \\ &\leq L^n \left(\frac{L+1}{L}\right) \|\varepsilon_0\|, \forall n \geq 1 \end{aligned} \quad (7)$$

Likewise, the lower bound of the cycle consistency can be derived with the assumption that

$$\|g_\theta(f \circ h(z_1)) - g_\theta(f \circ h(z_2))\| \geq l\|z_1 - z_2\|, \forall z_1, z_2 \in N_y \qquad (8)$$

In the neighbor of $y$, the two constants $L$ and $l$ can be approximated by the largest and smallest singular values of the Jacobian matrix $\frac{\partial}{\partial y} g_\theta(f \circ h(y))$, respectively.

In our analysis, we consider two common cases: the sequence of the cycle outputs $\{y_n, n = 0,1,2,\cdots,N\}$ diverges or converges. When $\{y_n\}$ diverges, i.e., $l \geq 1$, the lower bound can be expressed in terms of $\|\varepsilon_0\|$:

$$\begin{aligned}\|\Delta y_n\| = \|y_n - y_{n-1}\| &\geq l^{n-1}\|y_1 - y_0\| \\ &\geq l^{n-1}|\|g_\theta(f \circ h(y + \varepsilon_0)) - y\| - \|\varepsilon_0\|| \\ &\geq l^n \left(\frac{l-1}{l}\right)\|\varepsilon_0\|. \ \forall n \geq 1 \end{aligned} \qquad (9)$$

On the other hand, for $1 > L > l > 0$, the cycle outputs $\{y_n\}$ will eventually converge, and in this case, the lower bound can be written as:

$$\begin{aligned}\|\Delta y_n\| &\geq l^{n-1}|\|g_\theta(f \circ h(y + \varepsilon_0)) - y\| - \|\varepsilon_0\|| \\ &\geq l^n \left(\frac{1-L}{l}\right)\|\varepsilon_0\|. \ \forall n \geq 1\end{aligned} \qquad (10)$$

Based on the exponential form of the upper and lower bounds in Eqs. 7, 9 and 10 as a function of $n$, we can fit $\|\Delta y_n\|$ to an exponential function of the cycle index $n$. Therefore, we used the following regression relationship to estimate the uncertainty of the neural network output from cycle consistency $\|\Delta y_n\|$, *without* any knowledge of the ground truth:

$$\|\Delta y_n\| = k_y^n \varepsilon_y(1 + e_{n,y}) + b_y, n = 1,2,\cdots,N \qquad (11)$$

Here $e_{n,y}$ represents the random errors in the regression model that capture the effect of any unmodeled variables. For simplicity, we further assume that each $e_{n,y}$ independently follows a normal distribution with a variance of $s^2$, i.e., $e_{n,y} \overset{i.i.d.}{\sim} N(0, s^2), \forall n = 1,2,\cdots,N$. The coefficients $k_y$, $\varepsilon_y$ and $b_y$ represent (and model) the robustness, uncertainty, and bias of $g_\theta$ inference, respectively. In compliance with the upper and lower bounds reported in Eqs. 7, 9 and 10, Eq. 11 indicates that the cycle consistency $\|\Delta y_n\|$ should exhibit an exponentially increasing or decaying trend over the cycle index $n$. Although the theoretical bounds in Eqs. 7, 9 and 10 require the unbiasedness of the trained neural network, i.e., $y = g_\theta(f \circ h(y)), \forall y \in Y$, the regression relationship in Eq. 11 relaxes this assumption and can be applied to even biased neural networks.

We can perform a similar regression to the cycle consistency in the input domain, i.e., $\|\Delta x_n\|$:

$$\|\Delta x_n\| = k_x^n \varepsilon_x (1 + e_{n,x}) + b_x, n = 2,3,\cdots, N+1 \qquad (12)$$

Note here that since $\|\Delta x_1\| = \|x_1 - x\|$ is directly affected by the noise in the original input (measurement) $x$, it is not included in the regression analysis and is left as a separate measure to be used. The maximum cycle number $N$ should be selected in view of the applicability of our assumptions: $y_n, n = 1,2,\cdots, N$ cannot exceed the neighbor of $y$ where Eqs. 5 and 8 hold. For each input image and a given $N$, we perform the cyclic inference to get two sequences $\{x, x_1, x_2, \cdots, x_{N+1}\}$ and $\{y_0, y_1, \cdots, y_N\}$. Then, five uncertainty and bias estimators, i.e., $\hat{\varepsilon}_x, \hat{\varepsilon}_y, \hat{b}_x, \hat{b}_y, \|\Delta x_1\|$, are obtained from the regression analysis based on Eqs. 11 and 12.

To demonstrate the effectiveness of our method, we validated it with two OOD input detection tasks on two common inverse imaging problems used as our testbeds, which are detailed in the next sub-sections. First, we tested our method on a corrupted input detection task with an image deblurring network. We quantified the classification accuracy on various input data corruption cases, including different noise levels and distributions. Second, we applied our method to REAL-ESRGAN[40], an image SR network and detected OOD test images of unseen object classes. These two inference tasks consider data and model uncertainty separately to comprehensively evaluate our method for quantifying neural network uncertainty caused by various factors. In both tasks, the classifier was trained on a combination of ID data and OOD data generated by injecting noise, and the trained classifier, using cycle consistency metrics, was shown to distinguish OOD data perturbed by various unseen factors.

## 2.2 Corrupted input detection for an image deblurring neural network

We used the GoPro dataset[41] and the DeepRFT[42] model to implement image deblurring. Figure 2(a) illustrates a sharp image of a typical field of view (FOV) and a randomly generated motion blur kernel (see the Methods). The trained DeepRFT (indicated by the red arrows) could recover the sharp image from a noiseless blurry input image $x$, as shown in Fig. 2(b). For each blurred test image ($x$), the recovered sharp image $y_0$ was passed through the deterministic physical forward model (gray arrows) to generate the predicted input image $x_1$. In this way, a total of $N = 5$ cycles were implemented, and two sequences of images were created in the input and target domains, i.e., $\{x, x_n, n = 1,2,\cdots, N + 1\}$ and $\{y_n, n = 0,1,2,\cdots, N\}$, respectively. In Fig. 2(c), the cycle consistency $\|\Delta y_n\|$ and $\|\Delta x_n\|$ versus the cycle number $n$ are plotted, and the corresponding regression curves are illustrated, corresponding to Eqs. 11 and 12, respectively. Both curves demonstrate a good fit to the data points with coefficients of determination ($R^2$) of 0.9994 and 0.9960 for $\|\Delta y_n\|$ and $\|\Delta x_n\|$, respectively. Figure 2(d) further shows the actual uncertainty $\|\varepsilon_0\|$ versus $\hat{\varepsilon}_y$ over 100 blurry test images of the GoPro dataset generated with the shown blur kernel. The resulting data points indicate a positive correlation between $\|\varepsilon_0\|$ and $\hat{\varepsilon}_y$, confirming the effectiveness of our uncertainty estimator.

To further demonstrate the uncertainty quantification of our method, we fitted a linear regression model based on the five uncertainty and bias estimators ($\hat{\varepsilon}_x, \hat{\varepsilon}_y, \hat{b}_x, \hat{b}_y, \|\Delta x_1\|$) to predict the value of $\|\varepsilon_0\|$. The linear regression model was first trained on ~900 Gaussian noise-corrupted blurry input images of the test scene in the GoPro dataset with a random noise level $\sigma_{Gauss}$ from 0 to 0.1, and then tested on ~400 salt-and-pepper (SNP) noise-corrupted blurry input images of the test scene with a random noise level $\sigma_{SNP}$ from 0 to 0.1 (see the Methods for details). Figure 2(e) illustrates the predicted and ground truth uncertainty values on the training and testing sets of the linear regression model. This linear regression model accurately predicted the uncertainty for each input image in the training set, resulting in $R^2 = 0.9724$; after its training, the linear model could further generalize to the testing set with SNP noise corruption, achieving $R^2 = 0.8073$ (see Fig. 2(e)).

Next, we implemented corrupted input image detection using the cycle-consistency-based uncertainty and bias estimators. This detection was performed through a binary XGBoost[43] classifier using 5 attributes $\hat{\varepsilon}_x, \hat{\varepsilon}_y, \hat{b}_x, \hat{b}_y, \|\Delta x_1\|$, which were calculated through the cyclic inference and regression analysis of each input image as shown in Fig. 2(b, c). The classifier was trained on a dataset combining 1,000 ID input images and 1,200 corrupted (OOD) input images with random motion blur kernels and took ~1 sec to train (see the Methods for details). The corrupted input images were generated by introducing Gaussian noise with a noise level of $\sigma_{Gauss} \geq 0.01$ into the noiseless blurry image. Another 100 sharp images and 5 random kernels (excluded from training) were used to generate testing images. Figure 3(a) visualizes corrupted input images under various Gaussian noise levels $\sigma_{Gauss}$ from 0.00 to 0.03 and their corresponding outputs. As expected, we observed significant errors and artifacts in the output images corresponding to the corrupted input images with relatively high Gaussian noise levels, emphasizing the importance of corrupted input detection. In Fig. 3(b), a random subset of ID and OOD training and testing data are projected into a two-dimensional space formed by $\|\Delta x_1\|$ and $\hat{\varepsilon}_y$. Overall, the ID and OOD data are spatially separated, demonstrating the feasibility of corrupted input detection using the cycle consistency-based uncertainty and bias estimators. Furthermore, Fig. 3(c) quantifies the accuracy of the same classifier on a balanced dataset of ID input images and corrupted input images with Gaussian noise levels from 0.01 to 0.10. Two other baseline methods, i.e., an XGBoost classifier without cyclic inference and a supervised ResNet-50, were trained and tested on the same datasets for comparison (see the Methods for details). The XGBoost baseline took a similar time as our method for training and tuning (~1 sec per model), while the ResNet-50 baseline took ~12 hours to converge. As a simple machine learning algorithm without the need for time-consuming training, our method based on cycle consistency metrics matches the accuracy of the supervised ResNet-50 baseline over various noise levels, and shows considerably higher accuracy than the XGBoost baseline. The accuracy drops at around $\sigma_{Gauss} = 0.01$ as the data distribution approaches the boundary between ID and OOD data. However, our method significantly outperforms the two other baselines on this boundary; for example, when $\sigma_{Gauss} = 0.009$, our method scores an accuracy of 0.736, while the XGBoost and ResNet-50 scored lower accuracies of 0.576 and 0.485, respectively. In Fig. 3(d), we depict the 5 classification attributes ($\hat{\varepsilon}_x$, $\hat{\varepsilon}_y$, $\hat{b}_x$, $\hat{b}_y$, $\|\Delta x_1\|$) and the actual uncertainty as a function of the noise level $\sigma_{Gauss}$. The uncertainty and bias estimators (blue and purple lines) exhibit similar trends with respect to the actual uncertainty (gray lines), further validating the effectiveness of our quantitative uncertainty estimation. The importance of each attribute is also visualized in Fig. 3(d). Among the five attributes, $\hat{\varepsilon}_y$ and $\|\Delta x_1\|$ contributed more to the classification results than the other attributes, scoring relatively higher importance scores of 0.37 and 0.22, respectively.

We also blindly tested the same corrupted input image classifier, trained with the Gaussian noised inputs, on images with SNP noise that it never saw before. A summary of its classification performance on these corrupted inputs with Gaussian and SNP noise and its quantitative comparison against the two baseline methods are presented in Table 1. When tested on the balanced dataset of ID and Gaussian noise-injected OOD input images ($\sigma_{Gauss} \in [0.01, 0.10]$), our method scored an accuracy of 0.850, equivalent to the ResNet-50 baseline (0.845) and much higher than the XGBoost baseline (0.600). Furthermore, on the balanced dataset of ID and SNP noise-injected OOD input images ($\sigma_{SNP} \in [0.01, 0.10]$), our method scored an average accuracy of 0.980, which is better than the ResNet-50 baseline (0.845). These results prove the strong generalization of the OOD data classifier, trained on limited data, to distinguish corrupted data using cycle consistency-based metrics.

**2.3 Out-of-distribution detection for an image super-resolution (SR) neural network**
To validate our method's effectiveness on a different inverse problem with different sources of OOD data, we also implemented our method on the image SR task. For image SR networks, a common source of OOD data is distribution shifts, e.g., test images from object classes/types unseen in the training. In such cases, the external generalization of a neural network refers to testing on data distributions unseen in the training process, while internal generalization refers to testing on the dataset from the same distribution as in the training. External generalization usually results in more notable inference errors.

For this SR task, we selected the REAL-ESRGAN[44] model as our testbed, which demonstrates a good performance with checkpoints optimized for various image classes, including anime images and natural images. We implemented the cyclic inference process similar to the earlier experiment reported in the previous section, where $4 \times$ average pooling formed the forward measurement function $f \circ h$, and the SR model maps low-resolution measurements back to the domain of the high-resolution images. We empirically adopted the maximum cycle number to be $N = 3$. Figure 4(a) visualizes the training pipeline of our method and the two baseline methods: we first prepared OOD data by introducing noise into the training images of the model, e.g., noisy anime images were used as OOD data in the training of the classifier for the anime image SR model (see the Methods for more details). Then, similar to the training process used in the previous experiment (Section 2.2), our method implemented the cyclic inference (highlighted by the blue dashed lines in Fig. 4a) and trained the XGBoost classifier (blue arrows) to distinguish the network outputs for the ID images and the OOD noise-injected images, whereas both ID and OOD images belong to the same object class. Although the trained classifier in our method never saw other distribution shifts during its training, it could generalize to and classify new OOD data from unseen distributions, e.g., other object classes. The training processes of the ResNet-50 (green arrows) and the XGBoost (orange arrows) baselines (without any cycle consistency metrics used) are also illustrated in Fig. 4(a), where the ResNet-50 baseline directly performed classification on $y_0$ and the XGBoost baseline relied on $\|\Delta x_1\|$ for its classification. Our method and the XGBoost baseline took ~0.6 sec for training and tuning, whereas the ResNet-50 baseline took ~12 hours to converge. Although the XGBoost classifier in our method has lower model complexity and requires significantly less training time than standard deep neural networks, the reported cycle consistency-based quantitative metrics enhanced the XGBoost classifier to surpass the performance of the ResNet-50 baseline. Figures 4(b, c, d) quantify the classification accuracies of our method as well as the ResNet-50 and XGBoost baseline models on three classes of test images: anime, microscopy and human face images. As reported in Fig. 4(d), when the anime image SR model is externally generalizing on images of unseen object classes (microscopy images and human face images), our method classifies OOD data with an accuracy of 0.971 on the microscopy image dataset and with an accuracy of 0.891 on the face image dataset. For the other two specialized models (microscopy and face image SR models), our method can achieve accuracies of 0.743 and 0.809, respectively, to identify the OOD data. On the contrary, the baseline classification methods not using cycle consistency metrics shown in Fig. 4(b, c) significantly underperform compared to our method and cannot distinguish OOD images from different classes. The XGBoost baseline without cyclic inference lacks a high-dimensional description of uncertainty, and therefore results in worse classification accuracy, as shown in Fig. 4(c). For the ResNet-50 baseline, the distribution shifts between the anime images and the microscopy and face images were more significant than that between the anime ID and OOD training images, such that the trained classifier on anime images could easily distinguish the other two image classes but remained confused on identifying ID anime images (see the first row of Fig. 4(b)). Due to the limited number of training samples, the ResNet-50 baseline also tended to overfit to the training image class and could not generalize to unseen distribution shifts, as indicated by the failure on the second and third rows of Fig. 4(b). In contrast, our method successfully utilized the cycle consistency metrics and generalized to unseen distribution shifts, as shown by the relatively high off-diagonal accuracy scores in Fig. 4(d). Moreover, the diagonal entries (highlighted by the gray dashed lines) of the accuracy heatmaps shown in Fig. 4(b, c, d) report the accuracy scores identifying ID data that the SR models generalized well. Our method achieves >0.85 accuracy for such cases, avoiding excessive false positives.

## 3  Discussion and Conclusions

We introduced a novel UQ approach to quantify the inference errors of neural networks used in inverse problems and detect unseen corrupted input and OOD data. First, we built forward-backward cycles by iteratively running the physical forward model and the trained neural network, and established the theoretical relationship of the cycle consistency metrics in the expression of uncertainty, bias, and robustness of the neural network inference. Then, we derived uncertainty estimators through a regression analysis of the cycle consistency, without knowing the ground truth data. Finally, our method was validated on two applications: corrupted input detection on image deblurring tasks and OOD data detection on image SR tasks. Different from most existing UQ methods, our approach is model-agnostic and can adapt to various deep learning models in inverse imaging problems without modifications to the model architecture, training, or testing. On the corrupted input detection task with the image deblurring model, our method scored an accuracy of up to 0.980 on unseen corrupted input distributions; on the OOD data detection task with the image SR models, our method achieved up to 0.971 accuracy on unseen OOD image classes. Compared to the ResNet-50 baseline, our method achieved superior performance when generalizing on unseen distribution shifts, as reported in Table 1; when trained on Gaussian noise-injected images, our method scored 0.980 accuracy on unseen SNP noise-injected images, while the ResNet-50 scored a lower accuracy of 0.845.

In addition, our UQ method does not require prior knowledge of the OOD data distribution. In our experiments, the binary classifier was trained on simulated OOD data generated by injecting random noise of specific noise distributions, whereas the trained classifier could generalize to unseen OOD distributions from unseen sources, including e.g., different noise levels (Fig. 3), noise types (Table 1), and object classes (Fig. 4).

Future research could potentially apply deep neural networks such as ResNet to substitute the XGBoost classifier used in our method and further improve the classification accuracy using the uncertainty and bias estimators or the cyclic inference outputs directly. However, compared to the deep learning-based classifiers requiring millions of trainable parameters and hours of training, such as the ResNet-50 baseline used in this work, our method utilized the cycle consistency-based uncertainty estimators to enhance the standard XGBoost classifier that entails considerably less model complexity and less training time ($\leq$ 1 sec per model), providing a simple but efficient tool for OOD detection. In contrast, it took us ~12 hours on average to train the ResNet-50 baseline models with ~25M trainable parameters used in our comparisons.

Our UQ method also has some limitations. First, our UQ method originates from estimating the norm between the network outputs and ground truths, which often varies largely from image to image as an unnormalized metric. This issue impedes consistent quantitative uncertainty estimation across multiple datasets and physical forward models. As shown in Fig. 2, for input images of the same scene whose foreground and background have similar subjects and styles, a simple linear regression model can be established on the cycle consistency-based uncertainty estimators to provide quantitative uncertainty estimation per input image. However, fitting such a linear regression model may not always be possible, especially on training datasets with significant variations in the image objects and styles. Additionally, our cycle consistency metrics may not be able to detect hallucinations and artifacts generated by advanced neural networks, which produce photorealistic images with almost identical distribution as the ground truth[45,46]. In these cases, human perception or a discriminator with prior knowledge about the objects could be used to identify artifacts and hallucinations in the outputs of these networks. Second, our UQ method evaluates the comprehensive influence of data and model uncertainty without differentiating them from each other. However, as revealed in Fig. 3(d), the cycle consistency-based uncertainty estimators have different correlations to the measurement noise, which might be utilized to separate data and model uncertainty. Third, prior knowledge about the physical forward model $f \circ h$ is required by our UQ method. In some practical applications, the forward model is often parameterized by specific variables $\phi$, e.g., the blur kernel and the camera distortion/aberrations, and the prior knowledge or measurement of such variables during the blind test stage could be unavailable. Nevertheless, the cycle consistency metrics and the uncertainty/bias estimators in our method can potentially adapt to such situations with proper modifications. For one thing, our derivation of the relationship between the cycle consistency and uncertainty/bias does not require the exact forward model $f \circ h(\cdot)$ with ground truth parameters $\phi_{GT}$. Therefore, using the parameters of the physical forward model in the training $\phi_{train}$ to approximate $\phi_{GT}$ could be a promising solution, which needs further investigation. Furthermore, some of the existing models have explored simultaneous estimation of the forward model parameters and the ground truth signal in inverse problems[47–49]. Integrating the estimated forward model parameters into the cycle consistency evaluation process could be another future improvement to our method.

**Methods**

**Training of OOD classifiers**

The uncertainty and bias estimators derived in Section 2.1 are used as the attributes for OOD data classification. We picked XGBoost[43] as our classifier for this task. XGBoost is a highly accurate implementation of gradient boosting that builds parallel trees and follows a level-wise strategy to iteratively train decision trees using the previous model's error residual. Before training, we randomly split 80% of the training data into the training set, and the remaining 20% into the validation set.

To find the optimal parameters for XGBoost, we first used the random grid cross-validation search to fit and test the model with parameters randomly chosen from a wide range. The resulting parameters gave an estimation of the optimal parameters. We then tuned each parameter one by one using a grid search to select the best model. The optimal classification threshold was selected to maximize the F1 score on the validation set.

Additionally, we established two baseline models for comparison. The XGBoost baseline without cyclic inference adapted the same XGBoost training and tuning processes and utilized $\|\Delta x_1\|$ as the classification attribute. The second baseline leveraged the ResNet-50 architecture[50] to perform binary classification on the direct output $y_0$ and learned on the same training and validation datasets as our method and the XGBoost baseline.

**Testbeds: networks and datasets**

For the image deblurring task, we used the DeepRFT[42] model trained on the GoPro dataset[41]. We selected 900 sharp images (9 scenes) from the GoPro dataset and randomly split them into a training set of 800 images (8 training scenes) and a test set of 100 images (1 test scene) for the classifier, where the test scene was strictly excluded from the training set. Meanwhile, we randomly generated 26 motion-blur kernels to blur the training set images, resulting in ~1K ID and ~1.2K OOD training images. Motion blur kernels convolved with sharp images to generate blurry inputs and Gaussian noise $n_{Gauss} \sim N(0, \sigma^2_{Gauss})$ with a random level $\sigma_{Gauss}$ from 0 to 0.1 were introduced. The same operation was implemented on the testing images with 5 different blur kernels. SNP noise with a random level of $\sigma_{SNP}$ from 0 to 0.1 was also introduced to the test images to generate another test set to assess the classifier's generalization to other types of noise during the testing stage. The level of SNP noise $\sigma_{SNP}$ is defined as the ratio of the number of pixels whose values are randomly turned to 0 or 1 to the total number of pixels within each image. For the DeepRFT model, we empirically regard the input images with Gaussian noise levels lower than 0.01 as the ID data, since the corresponding outputs are almost indistinguishable from the outputs of their noiseless counterparts, as shown in Fig. 3(a). All the input images with Gaussian or SNP noise levels higher than or equal to 0.01 are classified as OOD data.

In the SR task, we used pre-trained models of REAL-ESRGAN[44], a widely-used image SR neural network for image restoration applications, and used 3 different image datasets, anime name and image dataset, microscopy image dataset, and Flickr-Faces-HQ dataset[51]. The microscopy image dataset was captured by a benchtop brightfield microscope (Leica Biosystems Aperio AT2) equipped with a $40 \times$ objective on hematoxylin and eosin (H&E) stained human lung, breast, and kidney tissue sections (from existing, deidentified samples and data)[52]. Throughout this paper, we refer to these three datasets as anime dataset, microscopy dataset, and face dataset. Three ESRGAN models were separately optimized for the three image datasets. First, a public ESRGAN model optimized for anime images was used for the anime dataset. Following the recommended training setup, we also finetuned the pre-trained ESRGAN model on 15000 face images and 15000 microscopy images to generate two SR models specialized for the face and microscopy datasets, respectively. Test sets of anime, microscopy and face images were excluded from the corresponding training datasets, and each test set contains 100 images.

For the training of the OOD data classifier for each image SR model, 100 random pristine images were selected from the corresponding training set of the SR model and served as ID data; random noise was injected into the ID images to create OOD data. The injected noise was randomly chosen from Gaussian and SNP noise processes and set at a random noise level from 0.02 to 0.05 for each OOD image. Besides, Gaussian and SNP noise, with a random noise level at 0.005 or 0.01, were added to ID data for data augmentation. As a result, the OOD data classifier for each image SR model was trained on 300 ID and 400 OOD images.

**Algorithm implementation**

All the algorithms and neural networks were implemented using Python and PyTorch framework[53] on a computer with an Intel Xeon W-2195CPU @ 2.30GHZ, 256 GB RAM and four Nvidia GeForce RTX 2080 Ti GPUs. For example, in the image SR task, the classifier in our method takes ~0.6s for training, and a single cyclic inference process ($N = 3$) takes approximately 0.71s for a high-resolution image with 1024×1024 pixels from the face dataset on a single RTX 2080 Ti GPU. The XGBoost baseline took a similar training time and slightly shorter inference time compared to our method since it implemented the same XGBoost classification processes but skipped the cyclic inference. The ResNet-50 baseline was trained from scratch and stopped after 1000 epochs to avoid overfitting. Standard augmentation techniques, including random flipping and rotation, were applied. The training time for the ResNet-50 baseline was ~12 hours, and the inference time on a single $1024 \times 1024$ image is 0.015s on the same machine using one RTX 2080 Ti GPU.

**Figures and Tables**

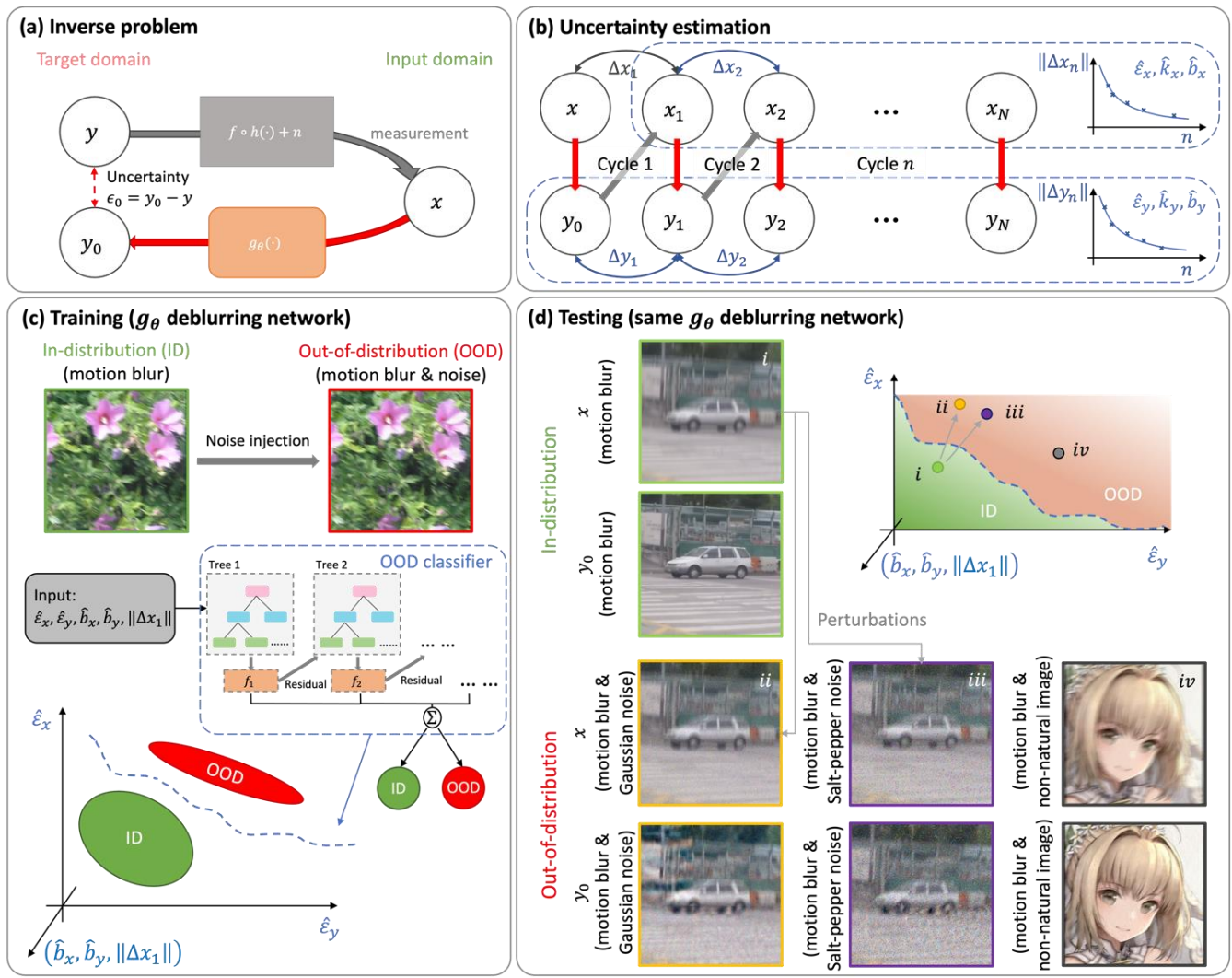

**Figure 1**. Uncertainty quantification (UQ) and out-of-distribution (OOD) detection using cycle consistency. (a) A diagram showcasing inverse problems. (b) Estimating uncertainty through regression analysis on cycle consistency. (c) Training of the OOD data classifier. (d) The trained classifier can distinguish unseen OOD data generated by various sources.

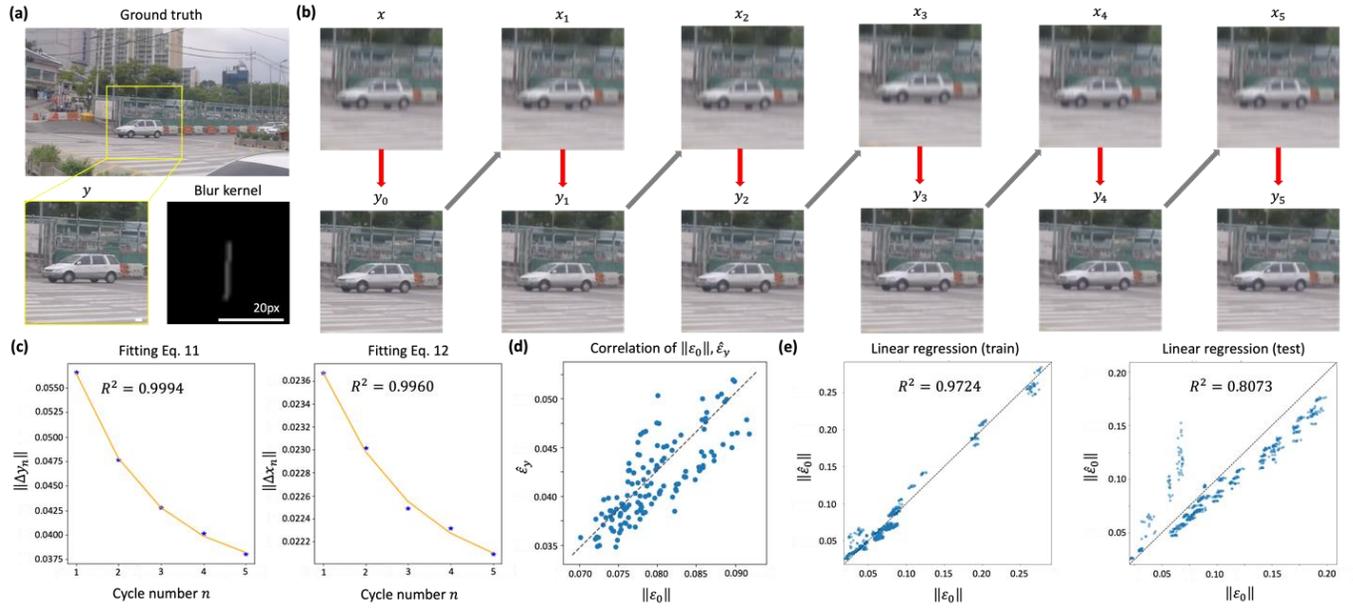

**Figure 2**. Uncertainty estimation using cycle consistency. (a) The sharp image (ground truth) and the motion blur kernel used. (b) The generated blurry image $x$ and the following cyclic inference, following Fig. 1b. (c) Regression analysis of the cycle consistency in (b). (d) Scatter plot of the uncertainty estimator and the actual uncertainty of 100 input images. (e) A linear regression model to predict the actual uncertainty. The linear regression model was first fitted on ~900 Gaussian noise-injected input images and then tested on ~400 SNP noise-injected input images.

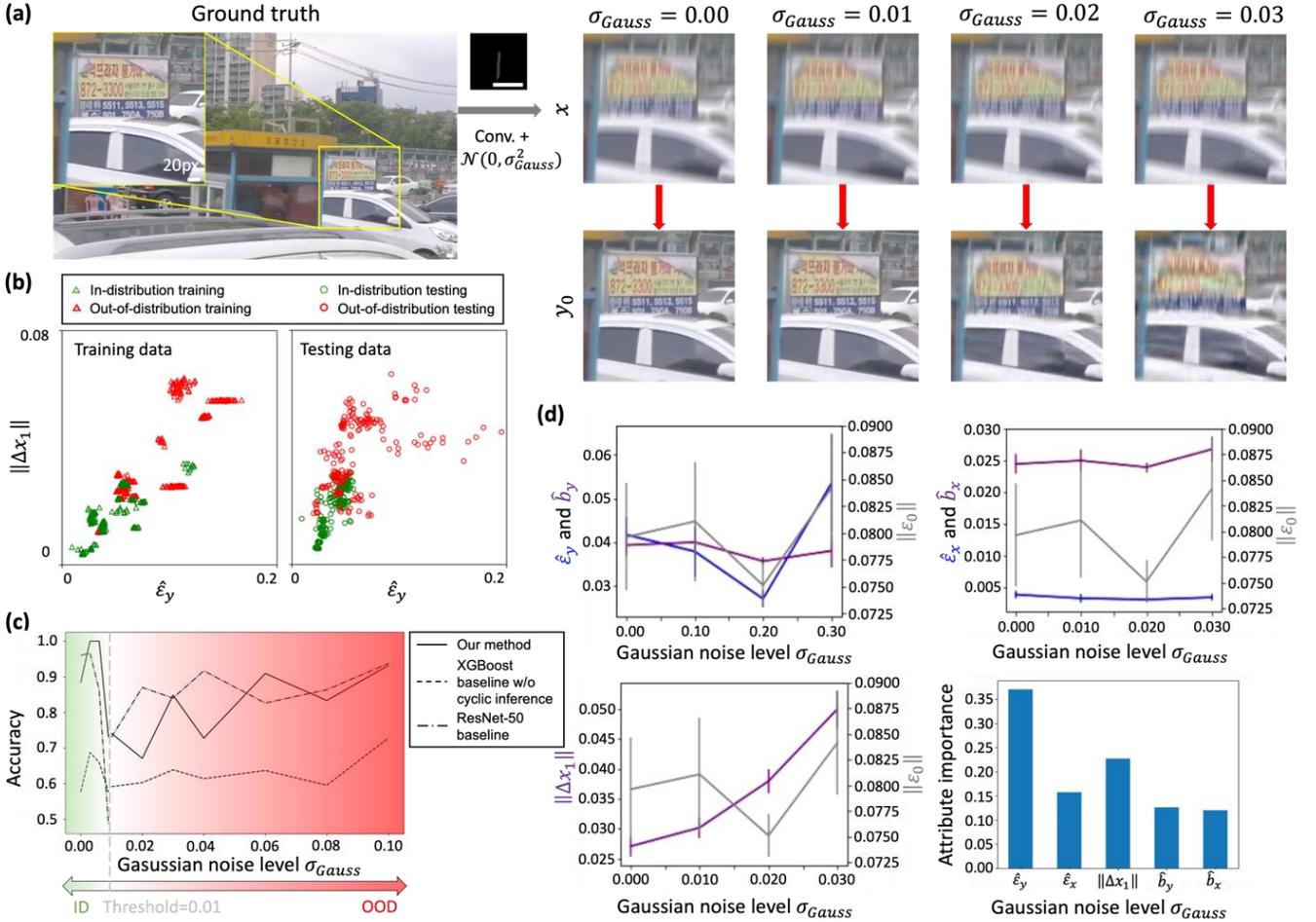

**Figure 3**. Corrupted input image detection using cycle consistency-based uncertainty and bias estimators. (a) Left: the sharp image (ground truth) and the motion blur kernel used. Right: the Gaussian noise-corrupted input images and the corresponding outputs. (b) Projection of the ID and OOD training and testing data on the two-dimensional space formed by the uncertainty estimators. (c) Detection accuracy of our method and two baseline methods on corrupted input images under various noise levels $\sigma_{Gauss}$. The classifiers were trained with ID images with $\sigma_{Gauss} < 0.01$ and OOD images with $\sigma_{Gauss} \geq 0.01$. (d) The estimated and actual uncertainty under various noise levels, and the importance of each attribute for ID vs. OOD classification. Scale bar: 20 pixels.

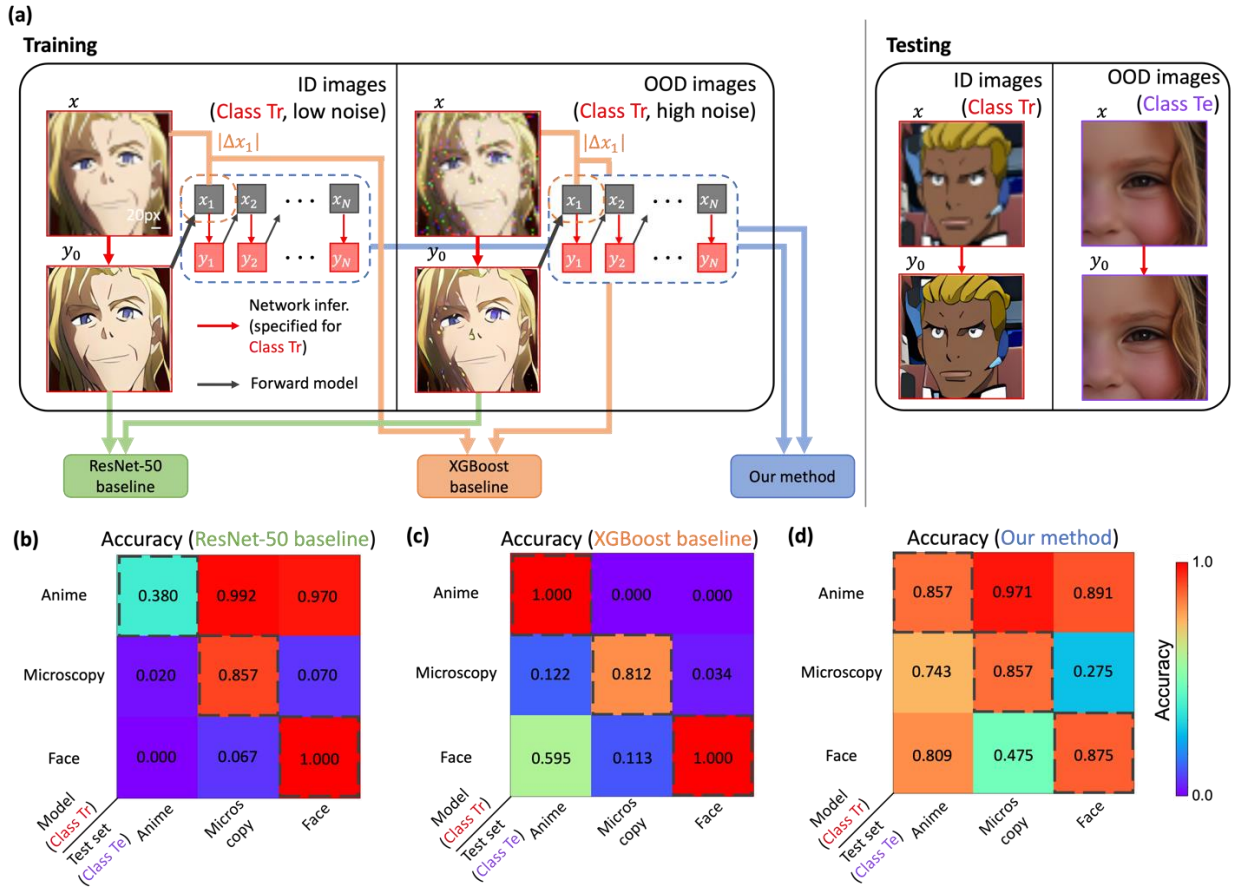

**Figure 4**. (a) Training pipelines of our method and the baseline methods (XGBoost and ResNet-50) for image SR models specialized for different image classes. The training OOD images were synthesized by injecting noise into the training images of each image SR model. After training, the classifiers were tested to distinguish ID images of the same class as training and OOD images of different classes. (b, c, d) OOD detection accuracy of ResNet-50, XGBoost and our method, respectively. For each method, a classifier was trained for each of the three image SR models specialized for anime, microscopy, and face images and tested on all three test sets. Diagonal entries highlighted by gray dashed lines represent the detection accuracy of ID images. Scale bar: 20 pixels.

**Table 1.** ID vs. OOD classification performance of our method and the baseline methods (XGBoost and ResNet-50). For each method, the classifier was trained on Gaussian noise-corrupted input images, and then tested on corrupted input images with Gaussian and SNP noise under various noise levels. The accuracy was calculated on balanced datasets with the same numbers of ID and OOD images. AP: average precision, AUC (ROC): area under the curve (receiver operating characteristic curve).

| Noise distribution | | **Gaussian Noise** | | | **SNP Noise** | | |
|---|---|---|---|---|---|---|---|
| Noise Level | | 0.01 | 0.02 | [0.01,0.10] | 0.01 | 0.02 | [0.01,0.10] |
| Our method | Accuracy | **0.805** | 0.739 | **0.850** | **0.978** | **0.978** | **0.980** |
| | AUC (ROC) | **0.847** | 0.842 | 0.917 | **1.000** | **1.000** | **1.000** |
| | AP | 0.753 | 0.691 | 0.788 | **0.952** | 0.953 | **0.962** |
| XGBoost baseline w/o cyclic inference | Accuracy | 0.595 | 0.636 | 0.600 | 0.594 | 0.597 | 0.635 |
| | AUC (ROC) | 0.761 | 0.602 | 0.682 | 1.000 | 1.000 | 1.000 |
| | AP | 0.553 | 0.579 | 0.556 | 0.523 | 0.526 | 0.578 |
| ResNet-50 baseline | Accuracy | 0.765 | **0.865** | 0.845 | 0.765 | 0.865 | 0.845 |
| | AUC (ROC) | 0.838 | **0.975** | **0.952** | 0.838 | 0.975 | 0.952 |
| | AP | **0.764** | **0.966** | **0.944** | 0.764 | **0.966** | 0.944 |

# References

1. Nguyen, G. *et al.* Machine Learning and Deep Learning frameworks and libraries for large-scale data mining: a survey. *Artif. Intell. Rev.* **52**, 77–124 (2019).

2. Otter, D. W., Medina, J. R. & Kalita, J. K. A Survey of the Usages of Deep Learning for Natural Language Processing. *IEEE Trans. Neural Netw. Learn. Syst.* **32**, 604–624 (2021).

3. Voulodimos, A., Doulamis, N., Doulamis, A. & Protopapadakis, E. Deep Learning for Computer Vision: A Brief Review. *Comput. Intell. Neurosci.* **2018**, 1–13 (2018).

4. Grigorescu, S., Trasnea, B., Cocias, T. & Macesanu, G. A survey of deep learning techniques for autonomous driving. *J. Field Robot.* **37**, 362–386 (2020).

5. Barbastathis, G., Ozcan, A. & Situ, G. On the use of deep learning for computational imaging. *Optica* **6**, 921–943 (2019).

6. Rivenson, Y. *et al.* Virtual histological staining of unlabelled tissue-autofluorescence images via deep learning. *Nat. Biomed. Eng.* **3**, 466–477 (2019).

7. Yang, J., Zhou, K., Li, Y. & Liu, Z. Generalized Out-of-Distribution Detection: A Survey. Preprint at http://arxiv.org/abs/2110.11334 (2022).

8. Hendrycks, D., Carlini, N., Schulman, J. & Steinhardt, J. Unsolved Problems in ML Safety. Preprint at http://arxiv.org/abs/2109.13916 (2022).

9. Goodfellow, I. J., Shlens, J. & Szegedy, C. Explaining and Harnessing Adversarial Examples. Preprint at http://arxiv.org/abs/1412.6572 (2015).

10. Huang, S., Papernot, N., Goodfellow, I., Duan, Y. & Abbeel, P. Adversarial Attacks on Neural Network Policies. Preprint at http://arxiv.org/abs/1702.02284 (2017).
22